\documentclass[]{fairmeta}

\usepackage{subcaption}

\usepackage{amsmath,amsthm,amssymb,bm, mathtools}
\usepackage{amsfonts}
\usepackage{thmtools} 
\usepackage{bbm}
\usepackage{algorithm}
\usepackage{algorithmic}

\usepackage{float}
\newfloat{algorithm}{t}{lop}
\usepackage{enumitem}

\usepackage{pifont}

\definecolor{RoyalBlue}{rgb}{0.25, 0.41, 0.88}

\newcommand{\cev}[1]{\reflectbox{\ensuremath{\vec{\reflectbox{\ensuremath{#1}}}}}}

\newenvironment{talign*}
 {\csname align*\endcsname}
 {\endalign}

\usepackage{xspace}

\DeclarePairedDelimiterX{\infdivx}[2]{(}{)}{%
  #1\;\delimsize\|\;#2%
}

\definecolor{cadetblue}{rgb}{0.37, 0.62, 0.63}
\definecolor{mygray}{gray}{0.95}
\definecolor{mediumpurple}{rgb}{0.46, 0.44, 0.68}
\definecolor{periwinkle}{rgb}{0.8, 0.8, 1.0}
\definecolor{purplemountainmajesty}{rgb}{0.59, 0.47, 0.71}

\makeatletter
\DeclareRobustCommand{\cev}[1]{%
  {\mathpalette\do@cev{#1}}%
}
\newcommand{\do@cev}[2]{%
  \vbox{\offinterlineskip
    \sbox\z@{$\m@th#1 x$}%
    \ialign{##\cr
      \hidewidth\reflectbox{$\m@th#1\vec{}\mkern4mu$}\hidewidth\cr
      \noalign{\kern-\ht\z@}
      $\m@th#1#2$\cr
    }%
  }%
}
\makeatother

\usepackage[noabbrev,nameinlink]{cleveref}
\theoremstyle{plain}  

\theoremstyle{definition}  

\title{M-TabNet: A Multi-Encoder Transformer Model for Predicting Neonatal Birth Weight from Multimodal Data}

\author[1,*]{Muhammad Mursil}
\author[1]{Hatem A. Rashwan}
\author[2]{Luis Santos-Calderon}
\author[3]{Pere Cavallé-Busquets}
\author[2]{Michelle M. Murphy}
\author[1]{Domenec Puig}

\affiliation[1]{Department of Computer Engineering and Mathematics, Universitat Rovira i Virgili (URV), 43007, Tarragona, Spain.}
\affiliation[2]{Unit of Preventive Medicine \& Biostatistics, Faculty of Medicine and Health Sciences, URV, CIBERObn ISCIII, 43201, Reus, Spain}
\affiliation[3]{Unit of Obstetrics \& Gynaecology, University Hospital Sant Joan, 43204 Reus, IISPV, CIBERObn ISCIII, Spain.}

\contribution[*]{Corresponding author}

\abstract{
Birth weight (BW) is a key indicator of neonatal health, with low birth weight (LBW) linked to increased mortality and morbidity. Early prediction of BW enables timely interventions; however, current methods like ultrasonography have limitations, including reduced accuracy before 20 weeks and operator-dependent variability. Existing models often neglect nutritional and genetic influences, focusing mainly on physiological and lifestyle factors. This study presents an attention-based transformer model with a multi-encoder architecture for early ($<12$ weeks) BW prediction. Our model effectively integrates diverse maternal data—physiological, lifestyle, nutritional, and genetic—addressing limitations seen in prior attention-based models such as TabNet. The model achieves a Mean Absolute Error (MAE) of 122 grams and an $R^{2}$ value of 0.94, demonstrating high predictive accuracy and interoperability with our in-house private dataset. Independent validation confirms generalizability (MAE: 105 grams, $R^{2}$: 0.95) with the IEEE children dataset. To enhance clinical utility, predicted BW is classified into low and normal categories, achieving a sensitivity of 97.55\% and a specificity of 94.48\%, facilitating early risk stratification. Model interpretability is reinforced through feature importance and SHAP analyses, highlighting significant influences of maternal age, tobacco exposure, and vitamin B12 status, with genetic factors playing a secondary role. Our results emphasize the potential of advanced deep learning models to improve early BW prediction, offering clinicians a robust, interpretable, and personalized tool for identifying pregnancies at risk and optimizing neonatal outcomes.
}

\begin{document}

\maketitle

\section{Introduction}
Birth weight (BW), defined as the first recorded weight of a neonate within the first hour after birth, is a critical indicator of neonatal health, significantly influencing survival rates, growth trajectories, and long-term developmental outcomes~\cite{islam2024determinants}. The World Health Organization (WHO) categorizes low birth weight (LBW) as less than 2,500 grams and macrosomia as exceeding 4,000 grams, both conditions associated with heightened risks of adverse health outcomes~\cite{shen2021prevalence}. LBW is linked to increased susceptibility to infections, higher incidences of chronic conditions such as cardiovascular diseases and diabetes in later life\cite{mursil2024maternal}, and a mortality risk 20 times greater than that of neonates with normal birth weight (NBW)~\cite{world2019unicef}. Conversely, macrosomia is often associated with complications during childbirth, including birth trauma and elevated rates of cesarean sections~\cite{niyi2024association}. Therefore, accurate early prediction of BW is essential for proactive identification of potential health risks, enabling timely medical interventions to improve neonatal outcomes and mitigate complications associated with abnormal BW.\par
Traditional BW prediction methods have predominantly relied on ultrasonography, a technique that, despite its widespread use, has notable limitations related to operator dependency, equipment quality, and accessibility—particularly in low-resource settings~\cite{sotiriadis2019systematic}\cite{sun2023liability}. Additionally, ultrasound becomes most reliable after 24 weeks of gestation, with optimal accuracy between 28 and 32 weeks. Before 20 weeks, its accuracy tends to be lower, making it less feasible for effective maternal health intervention \cite{dudley2005systematic}\cite{ewington2025accuracy}. Moreover, these methods often overlook the complex interplay of maternal physiological, nutritional, lifestyle, and genetic factors that significantly influence neonatal BW~\cite{ramiro2021maternal, gebremichael2024effect,santos2024folate, megaw2021higher, kobayashi2022gene}. This gap highlights the pressing need for predictive models to integrate multimodal maternal data, providing earlier, more precise, reliable, and interpretable predictions.\par

\begin{table*}[!ht]
\centering
\scriptsize 
\caption{SUMMARY OF DIFFERENT APPROACHES FOR BW PREDICTION AND HOW OUR WORK DIFFERS.} 
\label{approach}
\resizebox{\textwidth}{!}{ 
\begin{tabular}{|>{\raggedright\arraybackslash}m{2.5cm}|m{2.5cm}|m{2.5cm}|m{3cm}|m{4cm}|}
\hline
\textbf{Approach} & \textbf{Key Studies} & \textbf{Method} & \textbf{Limitations} & \textbf{How Our Work Differs} \\ \hline
\multirow{3}{2.5cm}{\textbf{Ultrasound-Based BW Prediction}} & Li et al. (2019)  \cite{li2019birth} & Regression models on 2D ultrasound & Accuracy declines before 20 weeks; operator-dependent & We predict BW $<$12 weeks gestation without ultrasound, using multimodal maternal data. \\ \cline{2-5}
 & Plotka et al. (2023) \cite{plotka2023deep} & Deep learning (Baby-Net) on ultrasound & Requires high-quality (HD) imaging and expertise & Our model does not depend on medical imaging, making it more accessible. \\ \cline{2-5}
 & Feng et al. (2019) \cite{feng2019fetal} & Hybrid ML model (SVM + DBN) & Limited generalizability & We use a transformer-based multi-encoder, improving generalization across datasets. \\ \hline
\multirow{3}{2.5cm}{\textbf{Demographic \& Clinical Data-Driven BW Prediction}} & Alabbad et al. (2024) \cite{alabbad2024birthweight} & ML models (ET, RF, SVR) on hospital datasets & Lacks interpretability and multimodal integration & We incorporate feature importance \& SHAP analysis, improving interpretability. \\ \cline{2-5}
 & Khan et al. (2022) \cite{khan2022infant} & RF model for BW classification & Does not leverage nutritional/genetic data & We include nutritional and genetic factors for a more holistic prediction. \\ \cline{2-5}
 & Ranjbar et al. (2023) \cite{ranjbar2023machine} & XGBoost on a large dataset & No feature selection insights & Our method ensures each modality contributes optimally using attention-based encoding. \\ \hline
 \multirow{3}{2.5cm}{\textbf{Multimodal Predictive Models (Bridging the Gap)}} & TabNet (Arik \& Pfister, 2021) \cite{arik2021tabnet}&Attention-based tabular learning &Struggles with multimodal data fusion  & Our multi-encoder architecture improves handling of heterogeneous features.\\ \cline{2-5}
 & Camargo et al. (2023) \cite{camargo2023multimodal} & Multiple kernel learning & Limited multimodal data integration and reply on HD imaging & Our model integrates multiple modalities simultaneously, enhancing cross-modal relationships for improved prediction. \\ \cline{2-5}
 \hline
\end{tabular}}
\end{table*}

In recent years, deep learning (DL) approaches, such as TabNet\cite{arik2021tabnet}, specifically designed for tabular data, have demonstrated remarkable improvements in predictive accuracy across various clinical applications. TabNet employs an attention mechanism to identify the most salient features at each decision-making step, applying feature masking to ensure sparsity and reduce computational complexity. However, despite its advantages, TabNet exhibits limitations in handling multimodal data effectively. Its attention mechanism may inadvertently prioritize features from a single data type, potentially neglecting the diversity of critical maternal information, thus impacting both predictive accuracy and interpretability.\par

To address these challenges, this study introduces a novel, transformer-based multi-encoder architecture to predict neonatal BW at an early gestational stage ($<$12 weeks). This approach effectively harnesses the rich, multimodal, and multidimensional nature of maternal data, encompassing physiological, nutritional, lifestyle, and genetic factors. Utilizing data from 730 mother-child dyads collected from two public hospitals in the Tarragona region, the proposed model achieves superior predictive performance, with a mean absolute error (MAE) of 122 grams and an r-squared (R²) value of 0.94. In addition, we classify the
predicted BW into low and normal BW categories with a sensitivity of 97. 55\% and a specificity of 94.48\%, demonstrating strong predictive performance in distinguishing neonatal BW categories. The model's interpretability is enhanced through feature importance, sensitivity analysis and Shapley Additive Explanations (SHAP), which elucidate the direction and magnitude of maternal factors' impacts on BW. 
This transformative approach represents a significant advancement in early BW prediction, offering a powerful tool for improving neonatal health outcomes through data-driven, personalized maternal care.

The primary contributions of this paper are as follows:\par
\begin{itemize}
     \item \textbf{Novel Transformer-Based Multi-Encoder Model:} We introduce a transformer-based multi-encoder architecture for early BW prediction ($<$12 GWs), effectively integrating multimodal maternal data (physiological, nutritional, lifestyle, and genetic) and overcoming limitations of models like TabNet by ensuring that each modality is appropriately represented in the prediction process.
     \item \textbf{Comprehensive Analysis of Underexplored Maternal Factors:} The study uniquely examines the combined influence of four key maternal factors, particularly investigating the relative contributions of underexplored maternal factors, such as nutrition and genotype, to neonatal BW prediction.
     \item \textbf{Clinically Interpretable Framework:} Our model ensures clinical relevance through SHAP, feature importance and sensitivity analysis, offering transparent insights into how maternal factors impact BW, enabling personalized care and early intervention strategies. 
\end{itemize}

\section{Related Work}

Accurate prediction of neonatal BW has long been a focus of maternal-fetal research, with traditional approaches relying on ultrasonography and, more recently, clinical data-driven predictive models. Table \ref{approach} provides an overview of the key studies on BW prediction and highlights how our work differs.

\subsection{Ultrasound-based BW prediction}
The table \ref{approach} highlights various studies using ultrasound-based techniques for predicting BW. Li et al. \cite{li2019birth} utilized 2D ultrasound data from 19,310 fetuses to develop regression models, but this approach may be affected by operator dependence and declining accuracy before 20 weeks of gestation. Plotka et al. \cite{plotka2023deep} leveraged deep learning with ultrasound, requiring high-quality imaging expertise. Although effective, these methods still rely on ultrasound technology, which limits their accessibility and applicability. Feng et al. \cite{feng2019fetal} proposed a hybrid machine learning (ML) model combining Support Vector Machine (SVM) and Deep Belief Networks (DBN) to predict BW based on ultrasound data. However, their model has limited generalizability, as it struggles to perform consistently across different datasets.

However, our model introduces a potential complementary tool to ultrasound by integrating multimodal maternal data, allowing for the prediction of BW before 12 weeks of gestation. Unlike ultrasound-based methods, which require specialized medical imaging, our approach is more accessible and applicable to a broader population, especially in settings where ultrasound may not be available. Additionally, by using a transformer-based multi-encoder, we improve generalization across diverse datasets, addressing the generalizability limitations seen in the recent key studies.

\subsection{Demographic and Clinical Data-Driven BW Prediction}

The previous reports primarily use demographic and clinical data to predict BW. Alabbad et al. \cite{alabbad2024birthweight} applied ML models, including Extra Trees (ET), Random Forest (RF), and Support Vector Regression (SVR), to hospital datasets but faced challenges related to multimodal integration and interpretability. Khan et al. \cite{khan2022infant} used the RF model for BW classification, yet their model does not consider genetic or nutritional data, limiting its predictive accuracy. Ranjbar et al. \cite{ranjbar2023machine} employed XGBoost on large datasets, but their approach lacked feature selection insights, which could optimize prediction.

In contrast, our model not only integrates demographic and clinical data but also incorporates underexplored maternal factors such as nutrition and genetic information. By doing so, we provide a more comprehensive prediction of BW, addressing the limitations faced by the studies mentioned above. We also improve the interpretability of our model using SHAP analysis, sensitivity analysis, and feature importance techniques, which are not widely used in traditional models.

\subsection{Bridging the Gap: Toward Multimodal Predictive Models}

TabNet \cite{arik2021tabnet} is a popular attention-based model for tabular data, widely used in predictive medicine for tasks like disease prediction and patient forecasting \cite{qi2024tab}. However, it struggles with multimodal data fusion, limiting its performance when integrating diverse data types. Camargo et al. \cite{camargo2023multimodal} use multiple kernel learning for BW prediction, but this approach processes each data modality separately, missing out on potential complex relationships between them.
Our transformer-based multi-encoder model improves multimodal integration by ensuring that each modality (e.g., nutritional, genetic, and lifestyle data) contributes optimally to BW prediction, overcoming the limitations of TabNet and recent studies. This ensures more accurate early predictions, considering the broader context of maternal health for personalized prenatal care.

\begin{table*}[t]
\centering
\scriptsize
\caption{COMPARATIVE STATISTICAL SUMMARY OF MATERNAL HEALTH INDICATORS IN THE REUS-TARRAGONA AND IEEE CHILDBIRTH DATASETS.}
\label{ReusData}
\resizebox{\linewidth}{!}{
\begin{tabular}{|p{0.2\linewidth}|p{0.3\linewidth}|p{0.2\linewidth}|p{0.3\linewidth}|}
\hline
\multicolumn{2}{|c|}{\textbf{Reus-Tarragona Dataset}} & \multicolumn{2}{c|}{\textbf{IEEE Childbirth Dataset}} \\
\hline
\textbf{Feature} & \textbf{Statistical Description} & \textbf{Feature} & \textbf{Statistical Description} \\
\hline
\multicolumn{4}{|c|}{\textbf{Physiological Factors}} \\
\hline
Maternal Age & $32 \pm 4.64$ & Maternal Age & $22 \pm 4.28$ \\
Previous Pregnancy & Yes: 391 (53.6\%), No: 339 (46.4\%) & Previous Pregnancy & $0.6 \pm 0.99$ \\
Gestational Weeks & $9.1 \pm 1.8$ GWs & Maternal Height & $142 \pm 17.24$ cm \\
Adverse Pregnancy History & Yes: 298 (40.8\%), No: 432 (59.2\%) & Fetal Sex & Male: 698 (51.7\%), Female: 652 (48.3\%) \\
Maternal BMI & $24.28 \pm 4.66$ kg/m$^2$ & Initial Systolic BP & $105.9 \pm 12.33$ \\
– & – & Initial Diastolic BP & $65.89 \pm 7.7$ \\
– & – & Final Systolic BP & $111.10 \pm 13.11$ \\
– & – & Final Diastolic BP & $70.61 \pm 8.57$ \\
– & – & Blood Group & A(+): 302 (22.4\%), A(-): 64 (4.7\%), B(+): 423 (31.3\%), B(-): 48 (3.5\%), AB(+): 168 (12.4\%), AB(-): 39 (2.8\%), O(+): 261 (19.3\%), O(-): 45 (3.4\%) \\
\hline
\multicolumn{4}{|c|}{\textbf{Nutritional Factors}} \\
\hline
Maternal Plasma Folate & $31.33 \pm 28.90$ nmol/L & Initial Hemoglobin & $10 \pm 1.05$ $\mu$mol/L \\
Maternal Vitamin B12 & $340.03 \pm 151.9$ pmol/L & Final Hemoglobin & $10.45 \pm 0.96$ $\mu$mol/L \\
Maternal Betaine & $15.6 \pm 3.83$ $\mu$mol/L & Blood Sugar & $100.66 \pm 11.48$ mg/dL \\
Maternal Choline & $8.07 \pm 1.73$ $\mu$mol/L & – & – \\
Maternal Anaemia (Hb $<$11 g/dL) & Yes: 12 (1.6\%), No: 718 (98.4\%) & – & – \\
\hline
\multicolumn{4}{|c|}{\textbf{Lifestyle Factors}} \\
\hline
Physical Activity & Low: 61.4\%, Medium: 37.6\%, High: 0.96\% & Socioeconomic Status & Below Poverty Line: 67.2\%, Above: 32.8\% \\
Tobacco Exposure & No: 73.7\%, Yes: 25.8\% & – & – \\
Socioeconomic Status & Lower: 11.6\%, Middle: 46.2\%, Higher: 42.2\% & – & – \\
Sun Exposure & Never: 29.6\%, Sporadic: 54.0\%, Regular: 16.4\% & – & – \\
\hline
\multicolumn{4}{|c|}{\textbf{Genetic Factors}} \\
\hline
MTHFR C677T & Wild: 33.9\%, Hetero+Homo: 66.1\% & – & – \\
MTRR A66G & Wild: 28.9\%, Hetero+Homo: 71.1\% & – & – \\
MTHFD1 105TC & Wild: 27.3\%, Hetero+Homo: 72.6\% & – & – \\
NOS7 T786C & Wild: 30.4\%, Hetero+Homo: 69.6\% & – & – \\
MTR A2756G & Wild: 64.4\%, Hetero+Homo: 35.6\% & – & – \\
\hline
\multicolumn{4}{|c|}{\textbf{Target Variable}} \\
\hline
Neonatal BW & $3230 \pm 470$ g & Neonatal BW & $2.7 \pm 0.43$ g \\
\hline
\end{tabular}
}
\end{table*}
\section{Methodology}
\subsection{Mathematical Formulation of the Problem}

Let \( X_m = \{ X_{\text{phys}}, X_{\text{nut}}, X_{\text{lifestyle}}, X_{\text{genetic}} \} \) represent the multimodal input data consisting of physiological, nutritional, lifestyle, and genetic factors, respectively. For each modality \( m \in \{\text{phys}, \text{nut}, \text{lifestyle}, \text{genetic}\} \), we have \( X_m \in \mathbb{R}^{n \times d_m} \), where \( n \) is the number of samples, and \( d_m \) is the dimensionality of the modality-specific data.

Each modality \( X_m \) is passed through a respective attentive transformer-based encoder \( E_m \), where the encoder learns the important features from each category for BW prediction, \( Z_m = E_m(X_m) \).  
The output of the attentive transformer \( Z_m \in \mathbb{R}^{n \times d_e} \), where \( d_e \) is the dimensionality of the encoder output. Each encoder is then masked to focus on the relevant features, \( \tilde{Z}_m = \text{Mask}(Z_m) \). The masked features for each modality \( \tilde{Z}_m \) are passed through a feature transformer \( T_m \) to extract higher-level representations, \( \hat{Z}_m = T_m(\tilde{Z}_m) \in \mathbb{R}^{n \times d_f} \), where \( d_f \) is the dimensionality of the higher-level representation.

The output of the feature transformer of all modalities is fused into a joint representation, \( Z_{\text{joint}} = \text{Fusion}(\hat{Z}_{\text{phys}}, \hat{Z}_{\text{nut}}, \hat{Z}_{\text{lifestyle}}, \hat{Z}_{\text{genetic}}) \in \mathbb{R}^{n \times d_{\text{joint}}} \), where \( d_{\text{joint}} \) is the dimensionality of the joint representation after fusion; mainly in this work, all the features are concatenated.

Finally, the joint representation \( Z_{\text{joint}} \) is passed through a decoder \( f \) to predict early neonatal BW, \( \hat{y} = f(Z_{\text{joint}}) \in \mathbb{R} \). 
The predicted BW \( \hat{y} \) is then classified as:  

\[
\hat{c} =
\begin{cases} 
\text{LBW}, & \text{if } \hat{y} < 2500 \text{ grams} \\
\text{NBW}, & \text{otherwise}
\end{cases}
\]


\subsection{Dataset}
\subsubsection{Reus–Tarragona Birth Cohort}
This work is based on data from the Reus–Tarragona birth cohort, a prospective longitudinal study designed to track maternal and child health outcomes from early pregnancy through childhood. The cohort is managed by the Unit of Preventive Medicine and Biostatics at the Faculty of Medicine and Health Sciences, Universitat Rovira i Virgili (URV), in collaboration with the Departments of Obstetrics and Gynecology at Sant Joan University Hospital in Reus and Joan XXIII University Hospital in Tarragona. The cohort is registered at ClinicalTrials.gov under the identifier NCT01778205.
The analysis focuses on data collected during the early pregnancy phase at $<$12 GWs. Pregnant women were recruited based on the following criteria:\par
\textbf{Inclusion Criteria:} Women with a confirmed singleton pregnancy and a viable fetus less than 12 weeks pregnant at their first prenatal blood sample collection.\par
\textbf{Exclusion Criteria:} Women were excluded if they were on medication known to affect folate or cobalamin status, had chronic diseases, had undergone surgical procedures affecting nutritional status, or were carrying multiple pregnancies.\par
Out of 831 pregnant women who initially consented to participate, some pregnancies did not result in a live birth  due to factors such as miscarriage, stillbirth, pregnancy termination due to foetal abnormalities and others were or were no longer monitored due to transfer to other healthcare facilities or loss to follow-up. A total of 730 mother-infant pairs with complete data on neonatal BW were included in the final analysis. See Table \ref{ReusData} for more details. 
\subsubsection{IEEE Child Birth Weight Dataset}
We use the IEEE Child Birth Weight Dataset\cite{dvd4323222}, a meticulously curated collection of maternal and neonatal health parameters, to ensure the proposed model's robustness and generalizability. This dataset is crucial for epidemiological studies, predictive modelling, and healthcare interventions related to BW and its factors. Compiled from a well-monitored birth cohort study in Assam, India, the dataset integrates diverse demographic, physiological, and nutritional indicators, offering a holistic view of the maternal health landscape. Data collected under the stringent oversight of medical professionals ensures high reliability, making it a valuable asset for research in perinatal healthcare.\par
The dataset, as shown in Table \ref{ReusData}, spans 1,350 pregnant participants and encompasses a spectrum of maternal attributes and district-level variables that exhibit substantial correlations with neonatal BW. These parameters provide a structured framework for investigating prenatal risk factors and optimizing the identification of useful parameters to consider in the research of early-life health outcomes. The dataset's depth and granularity make it particularly suited for advanced analytical techniques, such as ML-based prediction models, statistical inferences, and policy-driven healthcare enhancements. 

\subsection{Data Preprocessing}
We applied Data preprocessing to improve the accuracy and consistency of predictive models. In our study, we first identified and removed outliers in the Reus-Tarragona and IEEE datasets, such as BWs (e.g., 890 g, 4470 g, 600 g, 1000 g, and 4500 g), using the Interquartile Range (IQR) method. Missing values in longitudinal pregnancy studies is inevitable and were imputed in our time-series Reus-Tarragona dataset using linear interpolation, while categorical missing data were filled with Random Forest imputation. Random Forest imputation was also used to fill in missing values in the IEEE childbirth dataset. We applied MinMax scaling to standardize the datasets, transforming all values into a range between 0 and 1, ensuring that each feature contributed equally to the model’s performance, and preventing any one variable from disproportionately influencing the results. Additionally, to address the class imbalance, we applied the Synthetic Minority Over-Sampling with Gaussian Noise (SMOGN) technique \cite{branco2017smogn} only to the training dataset. This preprocessing approach ensured a balanced, normalized, and robust dataset for predictive analysis. Figure 1 shows the distribution of BWs in the original data and after applying SMOGN to both datasets. SMOGN generates samples in the poorly represented regions of the datasets.

\begin{figure}[htbp]
    \centering
    \includegraphics[width=0.8\columnwidth]{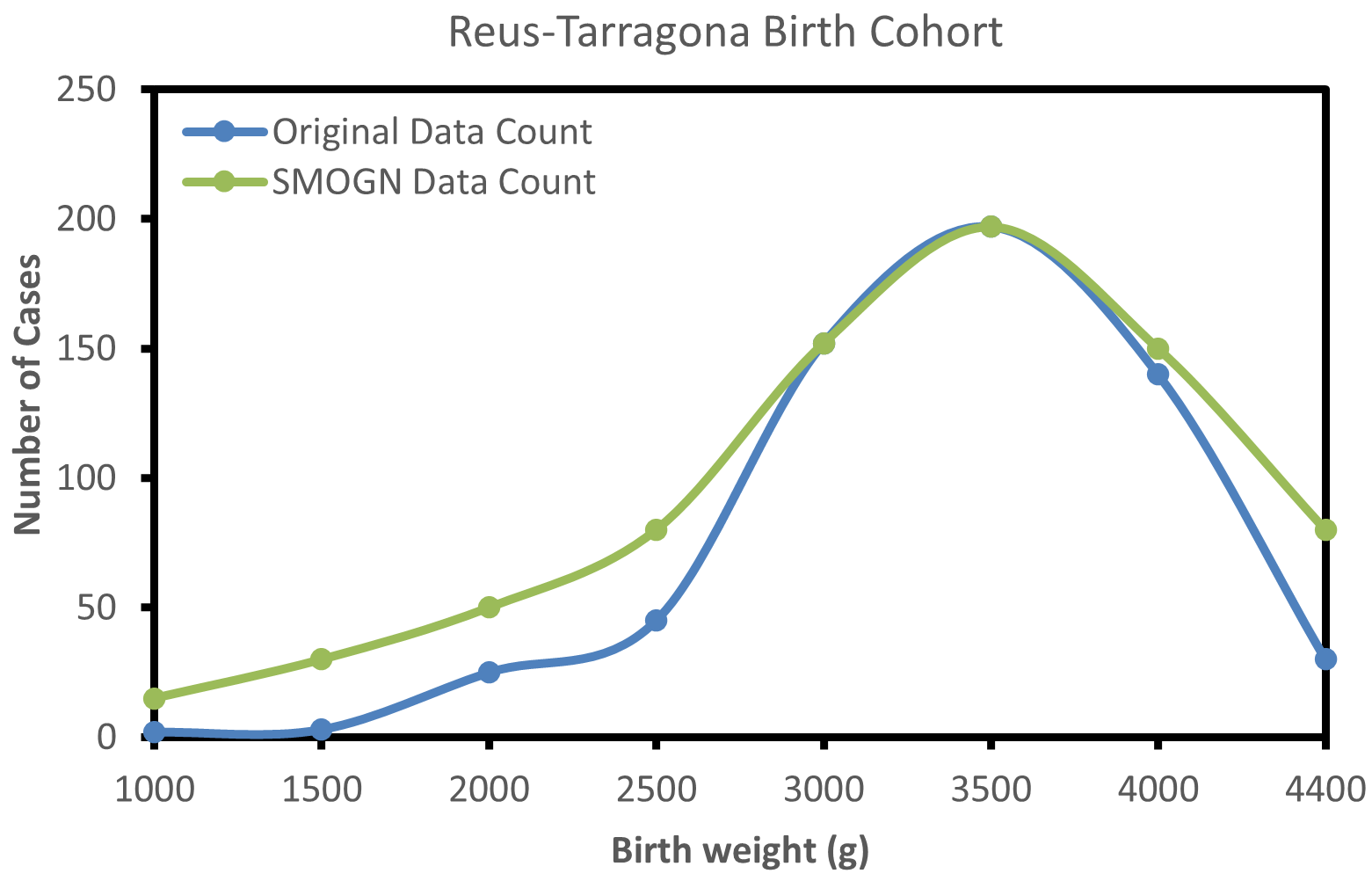} 
    \vskip\baselineskip
    \includegraphics[width=0.8\columnwidth]{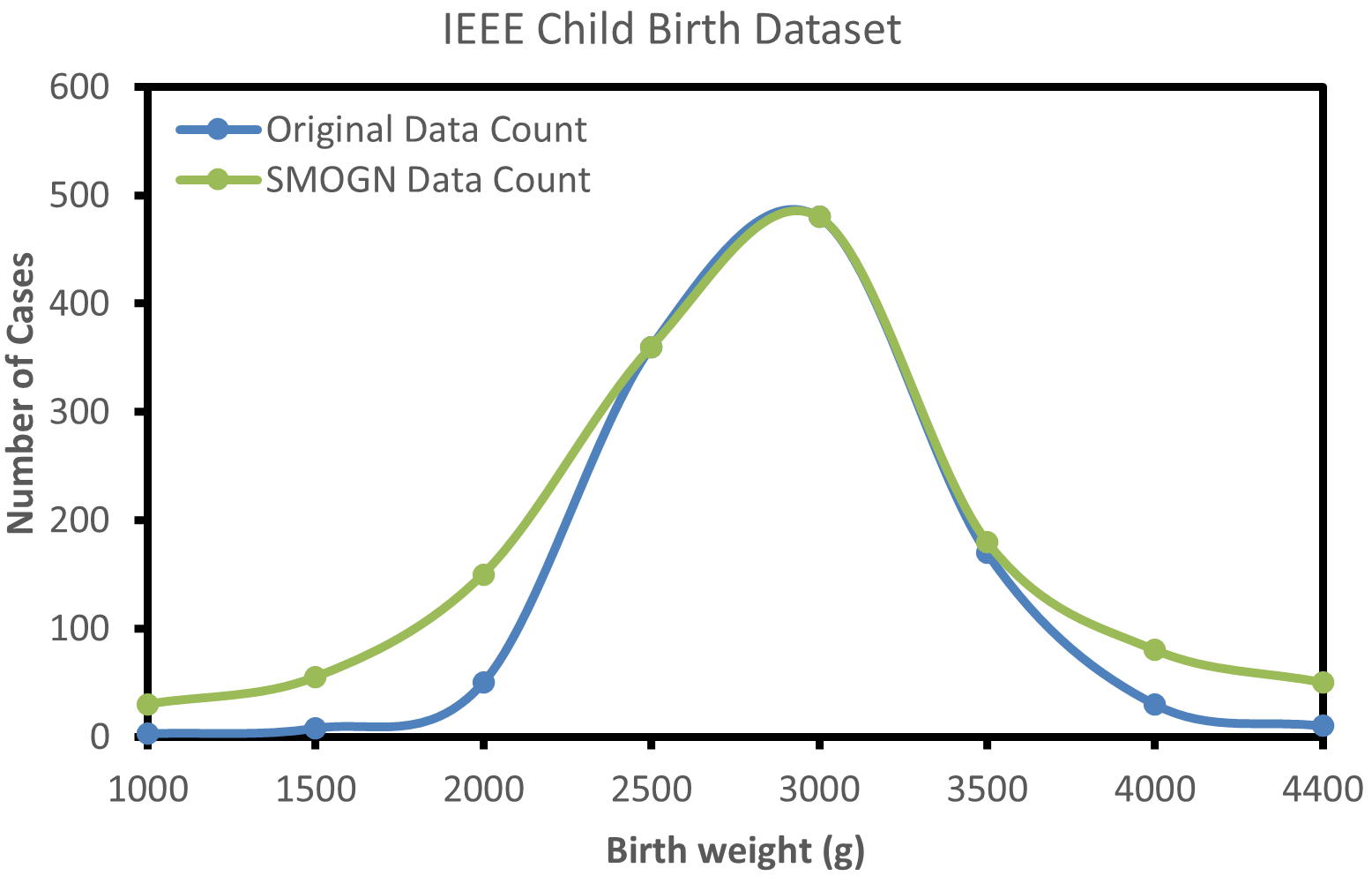} 
    \caption{Distribution of BWs in the original data and after applying SMOGN to the Reus-Tarragona Birth Cohort and IEEE Child Birth Dataset.}
    \label{fig1}
\end{figure}

\subsection{Transformer-based Multi-Encoder:}
\begin{figure*}[!ht]
    \centering    \includegraphics[width=1.0\textwidth]{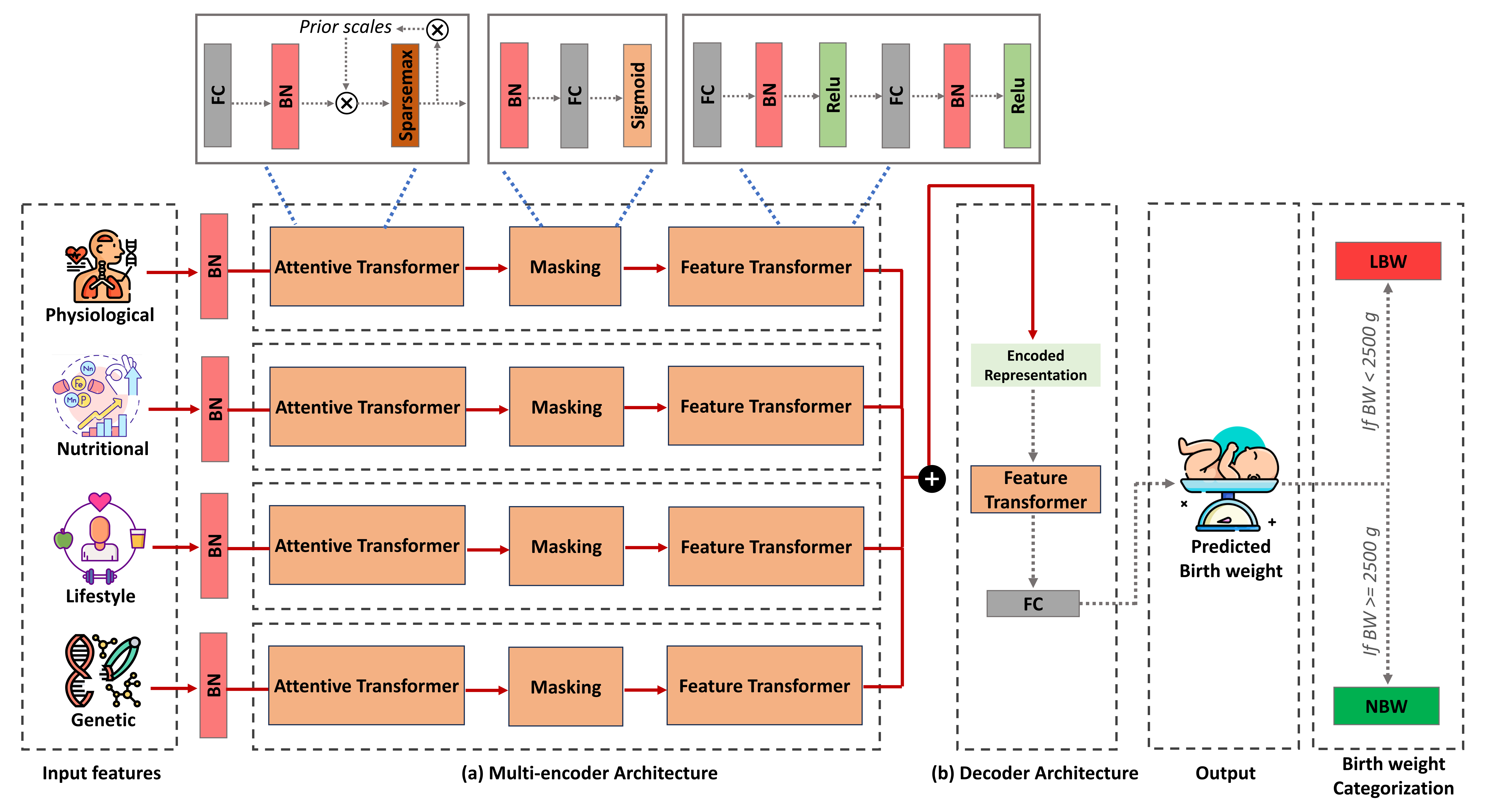} 
    \caption{Proposed Transformer-based Multi-encoder Architecture for Early Neonatal BW Prediction. The architecture integrates multiple encoders to process maternal nutritional, lifestyle, genetic, and health insights, facilitating more accurate prediction of BW and ultimately improving neonatal outcomes.}
    \label{fig2}
\end{figure*}

The Transformer-based multi-encoder proposed in this study is designed to overcome the limitations of TabNet, particularly in processing multimodal data within the domain of prenatal health. Inspired by TabNet’s \cite{arik2021tabnet} strength in capturing complex data relationships through sequential decision steps, the model utilizes attention mechanisms to prioritize the most relevant features, employing masks to maintain sparsity and reduce computational overhead. While TabNet performs well with single-modal data by processing selected features through a feature transformer (comprising fully connected layers) to extract intricate patterns and create higher-level representations, it struggles with precision and interpretability when dealing with multimodal data. This is because it may focus on features from only one modality, potentially overlooking critical information from others. To mitigate these issues, the proposed model enhances TabNet’s framework by incorporating transformer-based multiple encoders and multimodal feature fusion techniques tailored for multimodal data processing in prenatal health. This advancement retains TabNet’s attention-driven decision-making process. The proposed design significantly boosts the model’s capacity to handle diverse input modalities, enabling a more thorough analysis and improved retention of relevant features across various data sources. It causes enhanced interpretability and predictive performance, enabling easier integration into clinical decision support systems. The code for this work is publicly available on the
\href{https://github.com/mmursil5/M-TabNet}{\textcolor{red}{GitHub Repository}}. The model architecture is illustrated in Figure \ref{fig2} and consists of the following key components:\par

\subsubsection{Input Layer for Multimodal Data}

The model processes inputs from multiple modalities independently. Let the input for modality \( m \) be \( \mathbf{X}_m \in \mathbb{R}^{d_m} \), where \( \mathbf{X}_m \) is the input tensor for modality \( m \) and \( d_m \) is the number of features.

Each input modality undergoes batch normalization \(\mathbf{X}_m^{\text{norm}}\), normalizing the input tensor \( \mathbf{X}_m \) to have zero mean and unit variance:
\begin{equation}
\mathbf{X}_m^{\text{norm}} = \frac{\mathbf{X}_m - \mu_m}{\sigma_m},
\end{equation}
where \( \mu_m \) and \( \sigma_m \) are the mean and standard deviation of \( \mathbf{X}_m \) across the batch. Batch normalization stabilizes training by ensuring consistent feature distribution.

\subsubsection{Attentive Transformer for Feature Selection}
Each modality passes through its attentive transformer, which uses an attention mechanism to select the most relevant features of that modality. The attentive transformer employs \textit{Sparsemax} and a \textit{Prior scale} to enforce sparsity in feature selection, ensuring that only the most relevant features are highlighted. The attentive transformer has two key components:

\textbf{Attention Mechanism:} For modality \( m \), the attention mechanism computes:

\begin{equation}
\mathbf{H}_{\text{attn},m} = \text{softmax}\left(\frac{\mathbf{Q}_m \mathbf{K}_m^T}{\sqrt{d_k}}\right) \mathbf{V}_m,
\end{equation}
where \( \mathbf{Q}_m = \mathbf{h}_m \mathbf{W}_{Q,m} \), \( \mathbf{K}_m = \mathbf{h}_m \mathbf{W}_{K,m} \), \( \mathbf{V}_m = \mathbf{h}_m \mathbf{W}_{V,m} \) are the query, key, and value matrices for modality \( m \), and \( \mathbf{W}_{Q,m}, \mathbf{W}_{K,m}, \mathbf{W}_{V,m} \in \mathbb{R}^{d_h \times d_k} \) are learnable weight matrices for modality \( m \), with \( d_k \) being the dimension of the key vectors.
 The Softmax function is used in the attention mechanism for computing attention scores.

\textbf{Sparsemax and Prior Scale:}
\textit{Sparsemax} is a sparse activation function used in feature selection, particularly in the attentive transformer block, to enforce sparsity and focus on the most relevant features. This ensures that only the most relevant features are selected at each step.
The feature selection mask \( \mathbf{M}_m \in \mathbb{R}^{d_h \times d_h} \) for modality \( m \) is computed as:

\begin{equation}
\mathbf{M}_m = \text{Sparsemax}\left(\left( \prod_{j=1}^{\gamma} (\gamma - \mathbf{M}_{m,j}) \right) \cdot \mathbf{H}_{\text{attn},m} \right),
\end{equation}
where \( \prod_{j=1}^{\gamma} (\gamma - \mathbf{M}_{m,j}) \) is a prior scale term for modality \( m \), which aggregates how much each feature has been used in previous steps,

   \( \mathbf{H}_{\text{attn},m} \) is the output of the attention mechanism for modality \( m \), and \( \gamma \) is a relaxation parameter that controls the flexibility of feature reuse across steps.
The prior scale term is updated at each step to reflect the cumulative usage of features.

\subsubsection{Masking for Computational Efficiency}
The output of the attentive transformer for each modality \( m \) goes to the masking module. The mask \( \mathbf{M}_m \) essentially serves as a binary indicator that tells the model which features to consider at a given time. It ensures that irrelevant features do not influence the model's decision process at that stage.
The masked features for modality \( m \) are computed as:
\begin{equation}
\mathbf{H}_{\text{masked},m} = \mathbf{M}_m \cdot \mathbf{H}_{\text{attn},m},
\end{equation}
where 
    \( \mathbf{M}_m \) is the sparse mask matrix for modality \( m \),

\subsubsection{Feature Transformer}

After masking, the features for each modality \( m \) are passed through a feature transformer. The feature transformer processes the masked features to extract higher-level representations:

\begin{equation}
\mathbf{H}_{\text{transformed},m} = \text{FeatureTransformer}(\mathbf{H}_{\text{masked},m}),
\end{equation}
where 
    \( \mathbf{H}_{\text{transformed},m} \) is the transformed feature representation for modality \( m \),
    and \( \text{FeatureTransformer} \) is a neural network module that applies non-linear transformations to the input features.

\subsubsection{Multimodal Feature Fusion}
Following the feature transformer, the features from each modality are concatenated to form a comprehensive representation that captures all the relevant information from each data stream:

\begin{equation}
\mathbf{H}_{\text{fused}} = \text{Fusion}(\mathbf{H}_{\text{transformed},1}, \mathbf{H}_{\text{transformed},2}, \dots, \mathbf{H}_{\text{transformed},m}),
\end{equation}
where 
    \( \mathbf{H}_{\text{fused}} \) represents the concatenated features.

This concatenation step ensures that the final decision-making process can access a full range of modality-specific features and allows the model to retain the distinct contributions of each modality. This multimodal feature fusion process is crucial in improving the model’s performance, especially when the modalities provide diverse but complementary insights for BW prediction.

\subsubsection{Final Prediction Layer for BW Prediction}  
The fused features \( \mathbf{H}_{\text{fused}} \) are passed through a fully connected (FC) layer with a linear activation function to predict BW:  

\begin{equation}
\hat{y} = \mathbf{W}_{\text{out}} \mathbf{H}_{\text{fused}} + \mathbf{b}_{\text{out}},
\end{equation}

where  
    \( \mathbf{W}_{\text{out}} \in \mathbb{R}^{1 \times d_{\text{fused}}} \) is the weight matrix for the output layer,  
    \( \mathbf{b}_{\text{out}} \in \mathbb{R} \) is the bias term,  
    \( d_{\text{fused}} \) is the dimensionality of the fused feature vector. To enhance clinical relevance, the predicted BW \( \hat{y} \) is classified into distinct BW groups as follows:  

\begin{equation}
\hat{c} =
\begin{cases} 
\text{LBW}, & \text{if } \hat{y} < 2500 \text{ grams} \\
\text{NBW}, & \text{otherwise}
\end{cases}
\end{equation}



\subsection{Model Implementation, training and Evaluation}

The model was implemented in Python (version 3.12) and executed on PyCharm. The experiment was run on a Windows system with an Intel(R) Core(TM) i7-5930K CPU @ 3.50 GHz, 32 GB RAM. We employed 5-fold cross-validation to evaluate the proposed model's performance and generalization. 
We trained our model on the Reus-Tarragona birth cohort and IEEE dataset separately to assess the generalizability of the proposed model. To ensure optimal performance, we employed GridSearchCV for hyperparameter optimization, which allowed us to exhaustively search through a predefined set of hyperparameter values and identify the best-performing combination. Table 3 presents the optimal values of the hyperparameters that led to the highest model performance. We utilized MAE and $R^{2}$ as performance metrics for the evaluation. Additionally, sensitivity and specificity were used to assess the robustness of the proposed model in distinguishing BW categories, ensuring its reliability in identifying at-risk neonates.


\begin{table}[ht]
\centering
\caption{OPTIMAL HYPERPARAMETER VALUES FOR TRANSFORMER-BASED MULTI-ENCODER.}\label{tab2}
\begin{tabular}{|c|c|c|}
\hline
\textbf{Hyperparameter} & \textbf{Range} & \textbf{Optimal Value} \\ \hline
N\_steps & [3, 10] & 5 \\
Batch\_size & [32, 128] & 64 \\
Virtual\_Batch\_size & [16, 64] & 32 \\
Max\_epochs & [100, 500] & 200 \\
Optimizer\_params & [1e-1, 1e-3] & lr=2e-2 \\
Mask\_type & [sparsemax, entmax] & entmax \\
Gamma\_factor & [0.1, 1.0] & 0.9 \\ \hline
\end{tabular}
\end{table}

\subsection{Model Interpretability}
In this study, model interpretability was achieved using feature importance, sensitivity analysis, and SHAP analysis to gain insights into the factors influencing neonatal BW predictions. Feature importance was used to identify which maternal health factors had the most significant impact on the model’s predictions. A sensitivity analysis was performed by holding all features at their mean value and varying one feature at a time. Four representative values—minimum, mean-to-minimum median, mean-to-maximum median, and maximum—were used for each continuous feature, and the average changes in BW (\%) from the baseline were measured. For categorical features, sensitivity analysis was assessed on all values to evaluate their impact. This approach helped assess the influence of different features on the model’s output. SHAP analysis further provided a detailed understanding of how each feature contributed to individual predictions. 
Together, these analyses provided a comprehensive approach to interpreting the model, highlighting key insights into neonatal health.

\section{Results and Discussions}

\subsection{Model Predictive Analysis}
The predictive analysis was conducted on the Reus-Tarragona birth cohort dataset, using neonatal BW as the target label. Two evaluation stages were performed to compare different models: the original TabNet and the proposed M-TabNet with four variations (i.e., with all modalities, without genetic factors, without nutritional factors and without both genetic and nutritional factors), as shown in Table \ref{tab3}.
In the first stage, the TabNet model was trained on individual maternal factors. Physiological data, which is most likely available to clinicians, achieved the best performance among single modalities, with an MAE of 236.4 grams, an R² value of 0.8182, a sensitivity of 80.12\%, and a specificity of 76.68\%. Nutritional factors, which are less likely available, followed by an MAE of 280.1 g, an R² of 0.7512, a sensitivity of 68.78\%, and a specificity of 65.84\%. Lifestyle and genetic data—having most and least availability, respectively—performed worse, with MAE values of 314.9 g and 549.1 g, R² values of 0.7035 and 0.2805, sensitivities of 70.22\% and 52.67\%, and specificities of 68.46\% and 50.28\%, respectively. The low R² for genetic data indicates weak predictive power when used in isolation, suggesting its limited individual contribution to neonatal BW prediction. Then, all maternal factors (physiological, nutritional, lifestyle, and genetic) were integrated into the TabNet model, improving performance. The MAE dropped to 181.6 g, the R² increased to 0.8685, and the sensitivity and specificity improved to 86.85\% and 84.64\%, respectively.
In the second stage, the proposed M-TabNet model was introduced. This model separately processes each modality before integrating them, leading to a substantial performance improvement. M-TabNet achieved an MAE of 122.3 grams, an R² of 0.9432, a sensitivity of 97.55\%, and a specificity of 94.48\%, demonstrating its superior predictive ability. Further ablation studies revealed that removing genetic data slightly reduced performance (MAE = 151.9 g, R² = 0.9069, sensitivity = 93.54\%, specificity = 90.63\%), while eliminating nutritional data increased the MAE to 178.5 g (R² = 0.8736, sensitivity = 89.91\%, specificity = 87.65\%), highlighting its significant predictive contribution. Additionally, excluding both nutritional and genetic factors further deteriorated performance (MAE = 190.2 g, R² = 0.8698, sensitivity = 89.91\%, specificity = 87.65\%).
These results emphasize the complementary role of different maternal factors in neonatal BW prediction. While physiological data alone provides strong predictive power, integrating multiple factors, especially nutritional and lifestyle data, further improves accuracy. Moreover, M-TabNet's high sensitivity (97.55\%) and specificity (94.48\%) demonstrate its strong ability to distinguish between LBW and NBW neonates, making it highly effective in clinical settings. The superior performance of M-TabNet underscores the importance of multimodal data fusion in predictive modeling and establishes it as a complementary tool in clinical decision-making.

\begin{table*}[ht]
\centering
\scriptsize 
\caption{PERFORMANCE COMPARISON OF MODELS FOR NEONATAL BW PREDICTION USING MATERNAL FACTORS FROM THE REUS-TARRAGONA AND IEEE DATASETS.}

\label{tab3}
\resizebox{1.0\columnwidth}{!}{
\begin{tabular}{|c|c|c|c|c|c|c|}
\hline
\multicolumn{7}{|c|}{\textbf{Reus-Tarragona Birth Cohort}} \\ \hline
\textbf{Model} & \textbf{Maternal Factor} & \textbf{Availability to clinicians} & \textbf{MAE (g)} & \textbf{R$^2$} & \textbf{Sensitivity (\%)} & \textbf{Specificity (\%)} \\ \hline
TabNet & Physiological & Most Likely & 236.4 & 0.8182 & 80.12 & 76.68 \\
TabNet & Lifestyle & Most Likely & 314.9 & 0.7035 & 70.22 & 68.46 \\
TabNet & Nutritional & Less Likely & 280.1 & 0.7512 & 68.78 & 65.84 \\
TabNet & Genetic & Least Likely & 549.1 & 0.2805 & 52.67 & 50.28 \\
TabNet & Integrated all four modalities & - & 181.6 & 0.8635 & 86.85 & 84.64 \\
\textbf{M-TabNet} & \textbf{Integrated all four modalities} & - & \textbf{122.3} & \textbf{0.9432} & \textbf{97.55} & \textbf{94.48} \\
M-TabNet & without genetic & - & 151.9 & 0.9069 & 93.54 & 90.32 \\
M-TabNet & without nutritional & - & 178.5 & 0.8736 & 89.91 & 87.65 \\
M-TabNet & without nutritional and genetic & - & 190.2 & 0.8698 & 89.91 & 87.65 \\
\hline
\multicolumn{7}{|c|}{\textbf{IEEE Child Birth Data}} \\ \hline
M-TabNet & Integrated all three modalities & - & 105.4 & 0.9502 & 97.94 & 96.70 \\ \hline
\end{tabular}
}
\end{table*}

Figure 3(a) in the SM presents a scatter plot of the predicted versus actual BW values, demonstrating the model’s accuracy. The close alignment of points along the diagonal indicates that the model effectively captures neonatal BW variations. Figure 3(b) in SM shows the MAE distribution across the 5-fold cross-validation. The MAE remains consistently below 128 g across folds, confirming the model's stability, robustness and generalizability.

\begin{figure}[htbp]
    \centering
    \includegraphics[width=0.7\columnwidth]{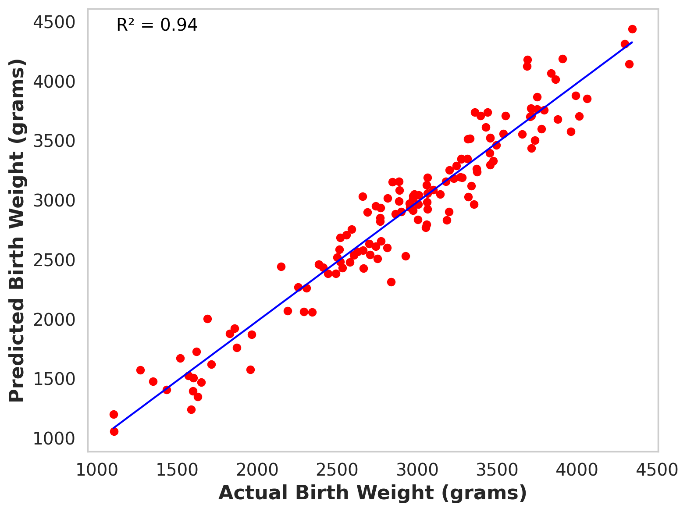} 
    \vskip\baselineskip
    \includegraphics[width=0.7\columnwidth]{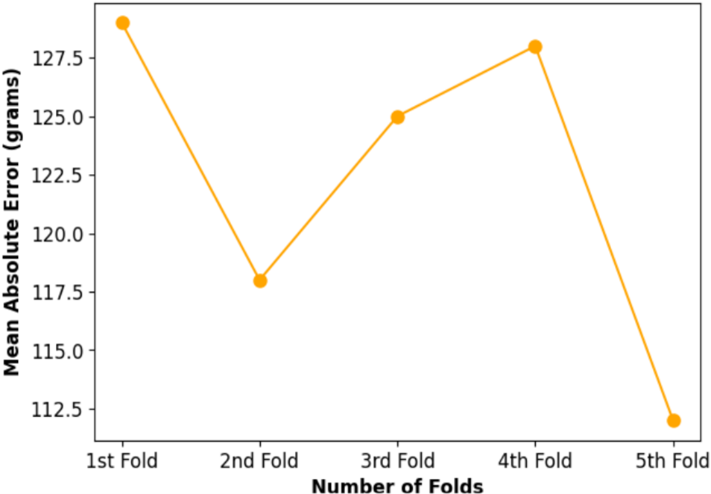} 
    \caption{Model Performance Evaluation: (a) Predicted vs. Actual BW across the R Line: The scatter plot shows the relationship between predicted and actual BW, with the blue line representing the linear regression fit. The R² value of 0.94 indicates a strong correlation between the predictions and actual values. (b) MAE during 5-fold cross-validation: The plot illustrates the changes in MAE across each fold, with the orange line highlighting the variation in performance as the model is trained and evaluated on different subsets of the data.}
    \label{fig1}
\end{figure}

\textbf{Model Generalization Analysis:} To evaluate the generalizability of the proposed model, we trained it from scratch on the IEEE childbirth dataset, which includes three maternal data modalities (physiological, nutritional, and lifestyle). The model achieved an MAE of 105.4 grams and an R² value of 0.95, maintaining high predictive accuracy despite being trained on a different dataset. This result demonstrates the model’s ability to generalize effectively across diverse maternal health data distributions, reinforcing its potential for real-world deployment in varied clinical settings.

\textbf{Paired t-test Analysis:} A paired t-test was performed on the results achieved from the Reus-Tarragona cohort to determine statistical significance. The proposed model significantly outperformed TabNet, with a t-statistic of 9.48 and a p-value of 0.0024 ($p<0.05$) for MAE. The R² comparison yielded a t-statistic of -9.79 and a p-value of 0.0022 ($p<0.05$), confirming that the proposed model provides a more accurate and explainable BW prediction.

\subsection{Ablation Studies}
To assess the contributions of individual model components, we conducted a series of ablation experiments as shown in Table \ref{tabAS}. The base model, which incorporates all four maternal data modalities (physiological, nutritional, lifestyle, and genetic) using attention mechanism, ReLU activation function, MinMax scaling, full training and concatenation for feature fusion, achieved an MAE of 122.3 g.

The attention mechanism played a critical role in feature selection and interaction modeling. Disabling attention led to the most significant performance drop, with the MAE increasing to 391.3 g (+269), demonstrating that the model heavily relies on attention to extract meaningful relationships between features. 
Other architectural and optimization choices also impacted performance. Switching the activation function from ReLU to GELU increased the MAE to 129.5 g (+7.2), indicating that while activation functions contribute to predictive accuracy, their impact is relatively minor. Feature scaling was also evaluated, with Z-score normalization increasing the MAE to 126.2 g (+3.9), suggesting that MinMax preserves relevant feature distributions more effectively. Early stopping provided a minor benefit, reducing the MAE to 119.8 g (-2.5), preventing overfitting while maintaining performance. Finally, we compared different fusion strategies. Replacing concatenation with aggregation increased the MAE to 140 g (+17.7), confirming that concatenation better retains cross-modal information, leading to superior performance.

\begin{table}[ht]
\centering
\scriptsize 
\caption{ABLATION STUDY ON MODEL COMPONENTS AND TECHNIQUES, SHOWING THE IMPACT OF MODALITY EXCLUSION, ATTENTION, ACTIVATION FUNCTIONS, SCALING, EARLY STOPPING, AND FUSION METHODS ON TEST MAE.}
\label{tabAS}
\resizebox{0.99\columnwidth}{!}{
\begin{tabular}{|p{0.55\linewidth}|>{\centering\arraybackslash}p{0.2\linewidth}|}
\hline
\textbf{Ablation Cases} & \textbf{MAE (Difference)} \\
\hline
M-TabNet (Base Model)  & 122.3 g \\
\hline
Without Attention Mechanism & 391.3 g (+269) \\
\hline
Activation Function (ReLU vs. GELU) & 129.5 g (+7.2) \\
\hline
Feature Scaling (MinMax vs. Z-score) & 126.2 g (+3.9) \\
\hline
Early Stopping vs. Full Training & 119.8 g (-2.5) \\
\hline
Fusion (Concatenation vs. Aggregation) & 140 g (+17.7) \\
\hline
\end{tabular}
}
\end{table}

\subsection{Model Interpretability Analysis}
\begin{figure}[!ht]  
\centering
    \includegraphics[width=0.8\columnwidth]{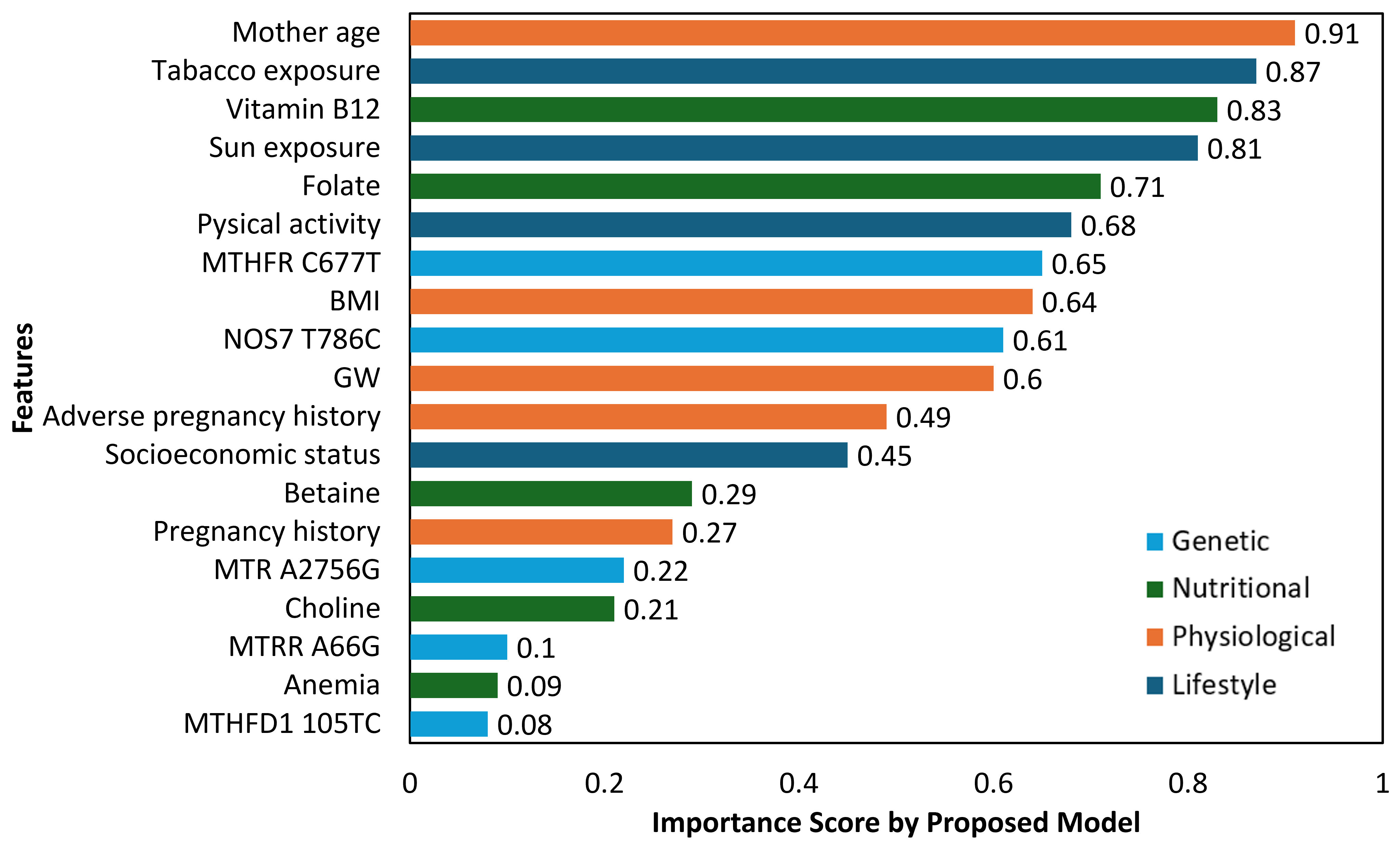}
\caption{Feature importance scores for neonatal BW prediction, highlighting age, tobacco use, and sun exposure as dominant factors. Anemia and the MTHFD1 105TC variant show minimal influence.}
\label{fig:figure4}
\end{figure}

To understand the decision-making process of the proposed model, we employed feature importance analysis, sensitivity analysis, and SHAP analysis. Figure \ref{fig:figure4} presents the feature importance scores, showing that maternal age (0.91), tobacco exposure (0.87), and vitamin B12 (0.83) are the most influential factors in predicting neonatal BW. Sun exposure (0.81) and folate (0.71) also contribute significantly, whereas anaemia and MTHFD1 105TC exhibit minimal influence. The relatively low impact of anaemia may be attributed to preventive maternal care protocols in the studied cohort.

\begin{figure}[!ht]
\centering
    \includegraphics[width=0.8\columnwidth]{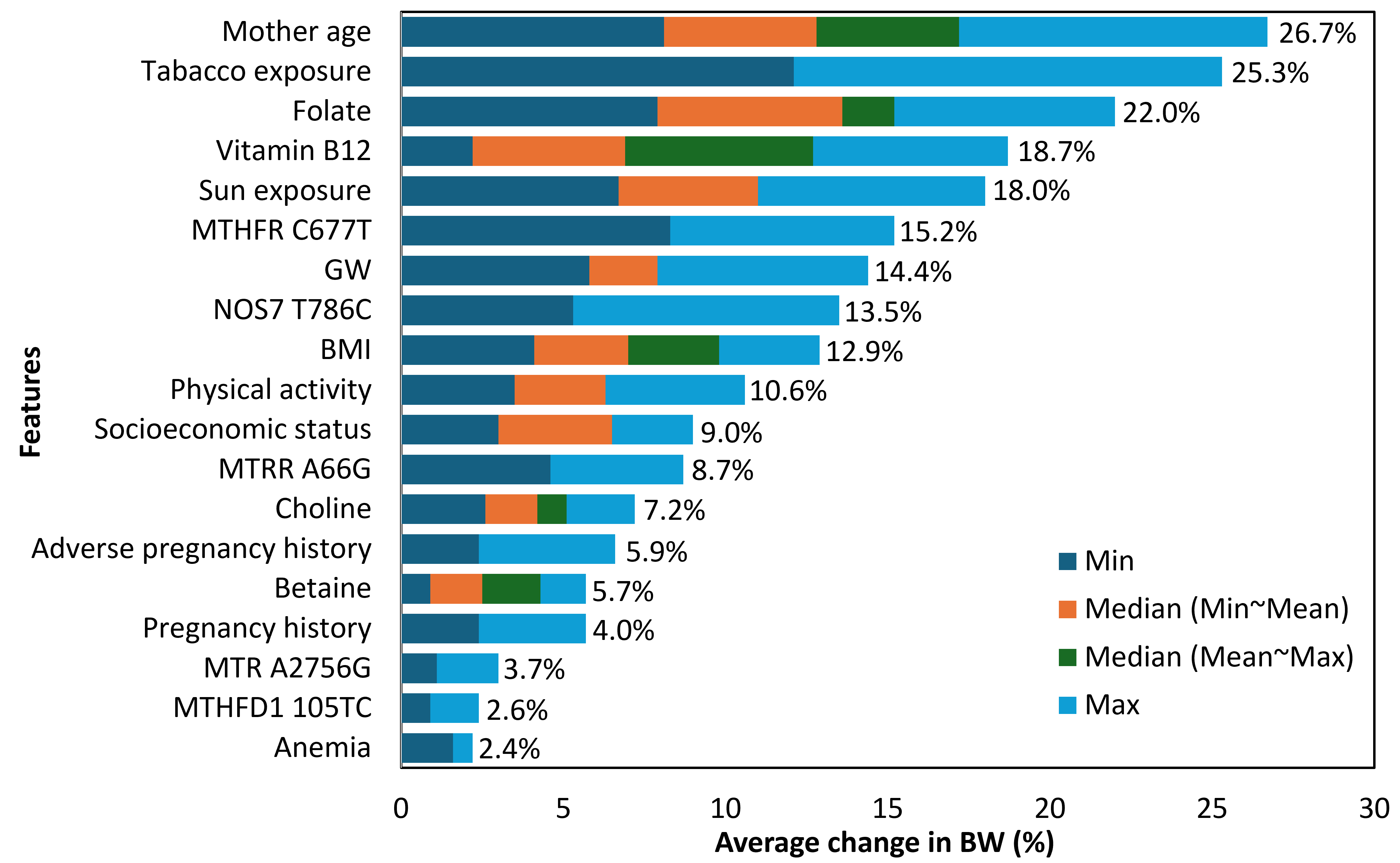}
\caption{Sensitivity analysis of maternal factors on neonatal BW shows age, tobacco use, and folate as key influencers, while anemia and MTHFD1 105TC had minimal impact.}
\label{fig:figure5}
\end{figure}

Figure \ref{fig:figure5} illustrates sensitivity analysis results, which quantify the effect of varying maternal factors on BW. Maternal age caused the most significant changes (26.7\% variation), followed by tobacco exposure (25.3\%), folate (22\%), and vitamin B12 (18.7\%). These results emphasize the role of lifestyle and nutritional factors in BW prediction, while anaemia and MTHFD1 105TC contributed minimally ($\leq$3\% variation).

\begin{figure}[!ht]
\centering
    \includegraphics[width=0.8\columnwidth]{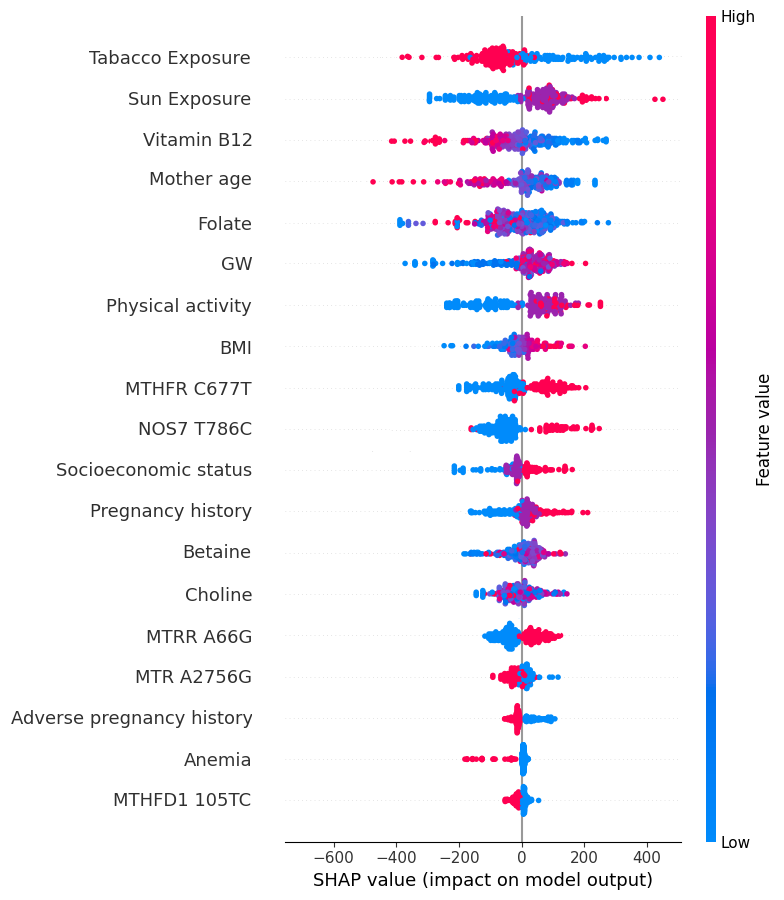}
\caption{SHAP analysis shows highest positive and negative impact of tobacco, sun exposure, vitamin B12, and maternal age in neonatal BW prediction.}
\label{fig:figure6}
\end{figure}

Figure \ref{fig:figure6} presents the SHAP analysis, which provides individualized feature impact on BW predictions. The ranking of predictors aligns with previous analyses, with tobacco use, sun exposure, vitamin B12, and maternal age emerging as key determinants. Tobacco exposure was associated with a reduction of approximately 350 grams in BW, while consistent sun exposure led to an increase of 410 grams. Vitamin B12 deficiency contributed to a BW increase of 250 grams, whereas maternal age was linked to a BW reduction of 430 grams. Anaemia and MTHFD1 105TC showed minimal impact.

The combined findings from these analyses reinforce the interpretability of our model, offering actionable insights for clinicians. The identification of maternal age, tobacco exposure, and vitamin B12 as dominant predictors suggests potential intervention strategies for mitigating risks associated with LBW. This interpretability allows healthcare professionals to assess risk factors early, personalize maternal care, and improve neonatal health outcomes.

\subsection{Clinical Applications and Limitations}
The proposed model can serve as an effective decision-support tool for clinicians, providing early neonatal BW predictions even in clinical settings where ultrasonography is unavailable. Integrating diverse maternal factors enables early identification of at-risk pregnancies, facilitating timely interventions such as nutritional support and lifestyle modifications. A potential application of this model includes a mobile or web-based platform that allows expectant mothers to monitor risk factors and receive personalized recommendations.

Despite its advantages, this study has some limitations. The relatively small sample size may limit its ability to detect subtle associations, and potential biases due to population stratification or confounding factors could affect model generalizability. Further validation on larger and more diverse cohorts is essential to ensure broader clinical applicability and enhance the robustness of the proposed approach. Gestational age at birth and fetal sex are well-established predictors of BW. However, these parameters are typically unavailable until the late or end of pregnancy, limiting their practical use in early prediction models. To assess their impact, we trained our models with completed gestational weeks at birth and foetal sex. While this inclusion improved predictive performance (MAE = 107 g, R² = 0.9477), it did not significantly alter the ranking of top feature importances, which remained strong predictors of BW. This suggests that despite their predictive value, the absence of these variables in early pregnancy does not substantially diminish our model's performance, reinforcing its robustness in real-world clinical settings.

\section{Conclusion}
This study introduces a novel transformer-based multi-encoder model for the early prediction of neonatal BW using a comprehensive set of multimodal maternal data encompassing physiological, nutritional, lifestyle, and genetic factors. By addressing the limitations of traditional methods, widely used ultrasonography scans, and existing deep learning models, such as TabNet, our proposed model significantly improves both the predictive accuracy and interpretability of BW predictions at an early gestational stage. The model's superior performance, demonstrated by a MAE of 122 grams and an R² value of 0.94, highlights its potential to make reliable early predictions, facilitating timely medical interventions.

Furthermore, the study emphasizes the underexplored role of maternal genotype and nutrition, in conventional models. The model's clinical relevance is enhanced by its transparent interpretability, achieved through sensitivity analysis and SHAP, which provide valuable insights into how variations in maternal factors affect neonatal health outcomes. The results suggest that maternal age, tobacco exposure, and vitamin B12 status are the most significant determinants of neonatal BW, with genetic factors playing a secondary role. 
This work represents a significant advancement in maternal health and child development, offering a promising tool for improving neonatal outcomes, particularly in settings where early intervention could mitigate complications associated with abnormal BWs. Future research may explore further model refinements and real-world clinical validations to expand its applicability and impact.

\section*{Acknowledgment}
Supported by grants from The Spanish Interministerial Science and Technology Committee (SAF2005-05096); The Carlos III Health Institute, National Scientific Research, Development and Technological Innovation Program Health Investigation Resources, cofinanced by The European Regional Development Fund (10/00335, 13/02500, 16/00506, 19/00844, CIBEROBn) The EPIBRAIN project (funded by the Joint Programming Initiative ‘A Healthy Diet for a Healthy Life’, JFA2 Nutrition and the Epigenome, Horizon2020 grant agreement number 696300 and The Spanish State Agency for Investigation PCI2018093098/AEI). Pere Virgili Health Research Institute (IISPV2010/21); Agency for Management of University and Research grants, Generalitat de Catalunya (Support to Research Groups: 2009-1237, 2014332) and Italfarmaco S.A., Spain. Predoctoral research fellowship from AGAUR et FI to Muhammad Mursil.\looseness=-1

\bibliographystyle{assets/plainnat}
\bibliography{paper}

\end{document}